\documentclass{article}

    \usepackage[preprint]{neurips_2024}

\usepackage[utf8]{inputenc} 
\usepackage[T1]{fontenc}    
\usepackage{hyperref}       
\usepackage{url}            
\usepackage{booktabs}       
\usepackage{amsfonts}       
\usepackage{nicefrac}       
\usepackage{microtype}      
\usepackage{xcolor}         
\usepackage{graphicx}       
\usepackage{amsmath}

\title{NusaMT-7B: Machine Translation for Low-Resource Indonesian Languages with Large Language Models}

\graphicspath{ {./images/} }

\author{%
William Tan \quad Kevin Zhu
\\
Algoverse AI Research\\
\texttt{william@tan.id, kevin@algoverse.us}
}

\begin{document}

\maketitle

\begin{abstract}
  Large Language Models (LLMs) have demonstrated exceptional promise in translation tasks for high-resource languages. However, their performance in low-resource languages is limited by the scarcity of both parallel and monolingual corpora, as well as the presence of noise. Consequently, such LLMs suffer with alignment and have lagged behind State-of-The-Art (SoTA) neural machine translation (NMT) models in these settings. This paper introduces NusaMT-7B, an LLM-based machine translation model for low-resource Indonesian languages, starting with Balinese and Minangkabau. Leveraging the pretrained LLaMA2-7B, our approach integrates continued pre-training on monolingual data, Supervised Fine-Tuning (SFT), self-learning, and an LLM-based data cleaner to reduce noise in parallel sentences. In the FLORES-200 multilingual translation benchmark, NusaMT-7B outperforms SoTA models in the spBLEU metric by up to +6.69 spBLEU in translations into Balinese and Minangkabau, but underperforms by up to -3.38 spBLEU in translations into higher-resource languages. Our results show that fine-tuned LLMs can enhance translation quality for low-resource languages, aiding in linguistic preservation and cross-cultural communication.
\end{abstract}

\section{Introduction}
Indonesia is home to 726 recorded regional languages, accounting for about 10\% of the world’s languages \citep{a2024_data}. While the official language, Indonesian, is spoken by 80.4\% of the population, a significant portion of these Indonesian speakers—about 70.9\%—are multilingual, often fluent in various regional languages \citep{a2024_indonesia}. However, predictions suggest that in 100 years, 90\% of these languages will either be extinct or on the verge of extinction \citep{osahitomiyaoka_2007_the}.

Machine translation systems have the potential to preserve endangered languages, serving as crucial tools for conservation efforts and fostering cross-cultural communication. However, low-resource languages, by definition, lack parallel corpora, which are crucial for traditional Neural Machine Translation (NMT) systems that often need millions of sentences for optimal performance. \citep{goyle_2023_neural}. While bitext mining datasets like CCMatrix \citep{schwenk_2021_ccmatrix} and NLLB can extract 1.4 million pairs, for instance, in the Minangkabau to English direction, a substantial portion of these datasets—98.7\% in this case—is plagued by noise and mismatched pairs.

Recent advancements in generative (decoder-only) Large Language Models (LLMs) have shown promise in machine translation. OpenAI’s GPT-3.5 and GPT-4 \citep{brown_2020_language}, along with advanced fine-tuned LLMs like ALMA-R \citep{xu_2024_contrastive} and Unbabel’s TowerInstruct \citep{alves_2024_tower}, have demonstrated remarkable performance in high-resource language translation, outperforming state-of-the-art (SoTA) models like NLLB-200 \citep{team_2022_no} and commercial translation products like Google Translate \citep{a2024_google} in the FLORES-200 translation benchmark. However, for low-resource languages, even very large models like GPT-4 lag significantly behind SoTA models like NLLB-200. Indeed, recent literature indicates that fine-tuned LLMs in machine translation are extremely sensitive to noisy data, easily picking up on erroneous biases and misalignments when noise is introduced \citep{zhu_2024_finetuning}. Thus, the primary bottleneck is likely not the inherent performance of LLMs, but the lack of high-quality, clean translation data typically found in low-resource language datasets.

Our proposed solution aims to bridge this gap. We introduce NusaMT-7B, a model focused on Indonesian low-resource languages, starting with Balinese and Minangkabau. Built on LLaMa2-7B \citep{touvron_2023_llama}, NusaMT-7B incorporates continued pre-training on non-English monolingual data, supervised fine-tuning, data preprocessing for cleaning parallel sentences, and synthetic data generation. We open-source NusaMT-7B on Huggingface\footnote{https://huggingface.co/xxx/xxx} and deploy a free translation web application\footnote{http://indonesiaku.com/} to showcase our model. We also release the training code\footnote{https://github.com/xxx/xxx} and compiled dataset\footnote{https://huggingface.co/xxx/xxx}. 

Our findings present three key takeaways:
\begin{enumerate}
\item Monolingual pre-training and a cleaner, smaller dataset both contribute to improved performance.
\item Backtranslation, a self-learning approach, boosts performance in LLM-based translation.
\item Our combined methods enable our model to outperform most SoTAs in the FLORES-200 benchmark for translation directions into low-resource languages.
\end{enumerate}

This paper includes a case study on the Balinese language to compare our proposed methods. We then extend our model to another low-resource Indonesian language, Minangkabau, and compare its performance against existing SoTA models.

\section{Related Work}

LLM-based machine translation has seen several innovative developments in low-resource languages. Previous research has focused on improving translation performance by incorporating human preference feedback \citep{jiao_2023_parrot,zhu_2024_a}. Other approaches have leveraged monolingual data for continued pretraining \citep{xu_2024_contrastive,alves_2024_tower}. Additionally, LLMs like GPT-4 have been used to clean noisy translation data, closely aligning with human cleaning and boosted performance when used to train NMT systems \citep{bolding_2023_ask}.

In the context of Indonesian low-resource languages, Komodo-7B-Instruct—a fine-tuned version of Komodo-7B-Base—has been trained on various tasks, including translation for languages like Balinese and Minangkabau. However, this model remains closed-source and has not been benchmarked against existing SoTA models in low-resource languages \citep{owen_2024_komodo}.

Our work, however, focuses on enhancing LLM performance through the integration of multiple paradigms, including parallel corpora cleaning, synthetic sentence pair generation, and monolingual pretraining. Together, we provide a comprehensive comparison against various SoTA models on the FLORES-200 benchmark in directions involving Balinese and Minangkabau.

\section{Proposed Method}
\subsection{Continued Pre-Training and SFT}
Most pretrained LLMs are unfamiliar with low-resource languages, making continued pre-training in the target low-resource language essential for teaching the LLM the linguistic principles of these previously unseen languages \citep{kuulmets_2024_teaching}. However, due to limited GPU resources required to pretrain on billions of tokens, we utilize Komodo-7B-base \citep{owen_2024_komodo}, a version of LLaMA2-7B further pretrained with Masked-Language Modeling (MLM) on 8.79 billion tokens. These tokens span a diverse set of multilingual corpora, including school textbooks and news articles from 11 regional Indonesian languages, such as Balinese and Minangkabau.

During SFT, we use the same translation prompt as ALMA \citep{xu_2024_a}, detailed in Appendix~\ref{translation_prompt}. The model is tasked with translating a sentence from a source language to a target language, with the loss computed only on the model’s generated tokens. Based on Xu et al.’s ablation study on training objectives for LLMs in machine translation, we employ Causal Language Modeling (CLM) loss for fine-tuning, which predicts the next word based only on the preceding context.


\subsection{LLM-based Data Preprocessor}

We propose an LLM data preprocessor tasked with (1) determining if two sentences share the same underlying meaning and, if so, (2) cleaning the parallel sentences to improve sentence alignment. We selected GPT-4o mini for this task due to its cost-efficiency and the superior performance in data preprocessing tasks demonstrated by its predecessor, GPT-4 \citep{openai_2023_gpt4}. Initially, we instruct the LLM on the tasks of data cleaning and aligning parallel sentences. We then use few-shot prompting to provide the LLM with representative examples of data cleaning, as shown in Appendix~\ref{datacleaner_prompt}. Batch prompting is used to process multiple parallel sentences in a single prompt to reduce overall token size.

\subsection{Backtranslation}
\label{sec:Backtranslation}
Backtranslation is a self-training method used to generate additional training data for SFT. It is a data-efficient method to augment new parallel sentence pairs and generate additional training data on more diverse linguistic structures and contexts. To generate synthetic sentence pairs from a source to a target language, we select high-quality sentences from monolingual datasets in the target language. After initially training our primary model with SFT, we run inference to translate the target monolingual data back into the source language. Subsequently, we apply our filtering methods and our LLM cleaner a second time to this new synthetic sentence pair to ensure proper alignment. Finally, we fine-tune the model to translate the source sentence back into the target sentence.

\section{Experiments}
\subsection{Data}

\begin{table}[h!]
\centering
\caption{Parallel sentence counts before and after LLM cleaning across datasets and language pairs including English (en), Indonesian (id), Balinese (ban), and Minangkabau (min)}
\label{dataset_counts}
\resizebox{\textwidth}{!}{%
\begin{tabular}{lcccccccc}
\toprule
\textbf{Dataset} & \multicolumn{4}{c}{\textbf{Before Cleaning}} & \multicolumn{4}{c}{\textbf{After Cleaning}} \\
\cmidrule(lr){2-5} \cmidrule(lr){6-9}
 & \textbf{ban $\leftrightarrow$ en} & \textbf{ban $\leftrightarrow$ id} & \textbf{min $\leftrightarrow$ en} & \textbf{min $\leftrightarrow$ id} & \textbf{ban $\leftrightarrow$ en} & \textbf{ban $\leftrightarrow$ id} & \textbf{min $\leftrightarrow$ en} & \textbf{min $\leftrightarrow$ id} \\
\midrule
NLLB Mined & 7.4k & 2.2k & 5.7k & 16.5k & 4.4k & 1.5k & 3.4k & 9.9k \\
NLLB SEED & 6.0k & 6.0k & 6.0k & 6.0k & 5.8k & 5.8k & 5.7k & 5.8k \\
BASAbaliWiki & 23.4k & 36.6k & 0 & 0 & 18.7k & 29.3k & 0 & 0 \\
Bible verses & 7.1k & 9.3k & 8.2k & 7.6k & 5.9k & 7.5k & 6.6k & 6.0k \\
NusaX & 0.9k & 1k & 1k & 0.9k & 0.8k & 0.8k & 0.9k & 0.7k \\
\midrule
\textbf{TOTAL} & \textbf{44.9k} & \textbf{55.2k} & \textbf{20.9k} & \textbf{31.0k} & \textbf{35.6k} & \textbf{44.9k} & \textbf{16.6k} & \textbf{22.4k} \\
\bottomrule
\end{tabular}
}
\end{table}

For our parallel data, we aggregated both human-annotated and automatically matched bitext datasets. We initially selected Balinese for our ablation study and subsequently expanded to Minangkabau using the best-performing method for additional benchmarking. Each low-resource language has four translation directions: to and from English and Indonesian. The number of parallel sentences for each dataset is shown in Table~\ref{dataset_counts}. It is important to note that all parallel sentences undergo a filtering pipeline as described in Appendix~\ref{filtering}.

We used AllenAI’s NLLB bitext dataset \citep{allenai_2024_nllb} (licensed under ODC-BY), sourced from metadata released by Meta AI as part of the NLLB project, as it is the largest dataset available for low-resource languages. Additionally, we used the human-annotated NLLB SEED dataset \citep{seed-23} (licensed under CC-BY-SA), which contains 6,062 sentences across English and multiple low-resource languages, including Balinese and Minangkabau.\footnote{Since SEED does not include Indonesian, and given the SoTA performance for en$\rightarrow$id, English SEED sentences were translated into Indonesian with the NLLB-3.3B model for additional bitext \citep{team_2022_no}.} We also extracted Bible verse bitexts from Alkitab.mobi \citep{AYT2018} (released under copyright for non-profit scholarly and personal use only), a collection of Bibles translated into regional Indonesian languages, where parallel sentences were generated automatically based on identical Bible line numbers. Finally, we used NusaX (licensed under CC-BY-SA), a parallel corpus annotated by Indonesian language experts across English and 10 Indonesian languages, including Balinese and Minangkabau \citep{indrawinata_2023_nusax}. For Balinese directions, we also sourced BASAbaliWiki (licensed under CC-BY-SA), a Balinese wiki containing articles with translations in Indonesian and English \citep{a2023_basabaliwiki}. In each article, we generated bitext by using LASER3 to find the nearest neighbors of each sentence and create possible sentence pairs, setting a similarity threshold of 0.7.

For monolingual data used in backtranslation, we aggregated sentences from Wikipedia dumps (CC BY-SA) and the Glot500 dataset, which was collected from other existing multilingual datasets (all of which we used were licensed under CC BY-NC or CC BY) \citep{imanigooghari_2023_glot500}.

We also report the parallel sentence counts before and after LLM cleaning in Table~\ref{dataset_counts}. Across all language pairs, there is a significant decrease in total parallel sentences—most notably from 31k to 22.4k sentences in the  min$\leftrightarrow$id pair. However, in human-annotated datasets like NLLB SEED and NusaX, minimal parallel sentences were filtered out, indicating that the LLM cleaner was proficient in retaining truly aligned sentence pairs.

\subsection{Training Setup}
In the initial SFT stage, we trained the model across all language directions simultaneously. To reduce GPU memory usage, we utilized Low-Rank Adaptation (LoRA) with a rank of 16, which reduces the number of trainable parameters to only 0.1\% (7.7 million from 7 billion parameters) \citep{hu_2021_lora}. We also used Deepspeed with ZeRO stage 2 offloading using bfloat16 to further optimize memory usage and enable multi-GPU training \citep{rajbhandari_2020_zero}. The dataset was randomly split into training, testing, and validation sets with 90\%, 5\% and 5\% splits respectively. Training was conducted over 3 epochs with a learning rate of 0.002 and a per-device batch size of 10. The best model weights were selected based on the lowest CLM loss on the validation set. For training, two Nvidia RTX 4090 GPUs were rented through the \href{https://vast.ai/}{vast.ai} cloud GPU platform. Training took 18 hours on our combined dataset, which includes Balinese and Minangkabau.

\subsection{A Balinese Case Study}
\label{bali}

\begin{table*}[h!]
\centering
\caption{spBLEU score comparison of the LLaMA2-7B SFT model with various enhancements, including monolingual pre-training (+ Mono), backtranslation (+ BT), and LLM cleaning (+ Cleaner)}
\vspace{0.5cm}

\label{tab:scores_ban}
\begin{tabular}{lccccc}
\toprule
\textbf{Models} & \textbf{ban $\rightarrow$ en} & \textbf{en $\rightarrow$ ban} & \textbf{ban $\rightarrow$ id} & \textbf{id $\rightarrow$ ban} \\
\midrule
Llama2-7B SFT & 27.63 & 13.94 & 27.90 & 13.68 \\
+ Mono & 31.28 & 18.92 & 28.75 & 20.11 \\
+ Mono + BT & 33.97 & 20.27 & 29.62 & 20.67 \\
+ Mono + Cleaner & 33.23 & 19.75 & 29.02 & 21.16 \\
+ Mono + Cleaner + BT & \textbf{35.42} & \textbf{22.15} & \textbf{31.56} & \textbf{22.95} \\
\bottomrule
\end{tabular}
\end{table*}

To study the effects of our proposed method, we compare fine-tuning the base LLaMA2-7B model with the addition of monolingual pre-training, LLM cleaning, and backtranslation methods. We present our findings in Table~\ref{tab:scores_ban}, using the spBLEU metric, which is the traditional BLEU metric applied over text tokenized by the FLORES-200 SentencePiece \citep{goyal_2022_the}.

The results indicate that the Komodo-7B-base model, with additional monolingual pre-training, achieves substantial gains over the base LLaMA2-7B model across all translation directions. Notably, we observe up to a 45\% increase in spBLEU in the Indonesian to Balinese direction. Additionally, we find that the LLM cleaning method alone raises spBLEU scores by an average of 5\%. This suggests that a reduced training size with reduced noise can indeed boost model performance, supporting the LIMA hypothesis. We also report a 4.7\% increase in spBLEU through backtranslation, demonstrating the LLM's capacity to continue learning through synthetically generated data. Furthermore, when applying both methods in conjunction, we observe an average performance increase of 13\% spBLEU over the Mono + SFT baseline.

\subsection{Benchmarking}

\textbf{Baselines.} We now evaluate and benchmark NusaMT-7B, our model with the LLM cleaner and backtranslation, additionally trained on Minangkabau language directions. First, we benchmark the SoTA NLLB-200 models, including the 3.3B, 1.3B, and the distilled 600M variant. Additionally, we benchmark the very large GPTs from OpenAI—GPT-3.5-turbo, GPT-4, and the latest GPT-4o—using zero-shot prompts.

\begin{table*}[h!]
\centering
\caption{spBLEU scores of NusaMT-7B compared against SoTA models (NLLB-600M, NLLB-1.3B, NLLB-3.3B) and large GPT models (GPT-3.5-turbo, GPT-4o, GPT-4)}
\vspace{0.5cm}

\label{tab:scores_benchmark}
\resizebox{\textwidth}{!}{%
\begin{tabular}{lcccccccc}
\toprule
\textbf{Models} & \textbf{ban $\rightarrow$ en} & \textbf{en $\rightarrow$ ban} & \textbf{ban $\rightarrow$ id} & \textbf{id $\rightarrow$ ban} & \textbf{min $\rightarrow$ en} & \textbf{en $\rightarrow$ min} & \textbf{min $\rightarrow$ id} & \textbf{id $\rightarrow$ min} \\
\midrule
GPT-3.5-turbo, zero-shot & 27.17 & 11.63 & 28.17 & 13.14 & 28.75 & 11.07 & 31.06 & 11.05 \\
GPT-4o, zero-shot & 27.11 & 11.45 & 27.89 & 13.08 & 28.63 & 11.00 & 31.27 & 11.00 \\
GPT-4, zero-shot & 27.20 & 11.59 & 28.41 & 13.24 & 28.51 & 10.99 & 31.00 & 10.93 \\
\cmidrule(lr){1-9}
NLLB-600M & 33.96 & 16.86 & 30.12 & 15.15 & 35.05 & 19.72 & 31.92 & 17.72 \\
NLLB-1.3B & 37.24 & 17.73 & 32.42 & 16.21 & 38.59 & 22.79 & 34.68 & 20.89 \\
NLLB-3.3B & \textbf{38.57} & 17.09 & \textbf{33.35} & 14.85 & \textbf{40.61} & \textbf{24.71} & \textbf{35.20} & 22.44 \\
\cmidrule(lr){1-9}
NusaMT-7B (Ours) & 35.42 & \textbf{22.15} & 31.56 & \textbf{22.95} & 37.23 & 24.32 & 34.29 & \textbf{23.27} \\
\bottomrule
\end{tabular}
}
\end{table*}

We report our benchmarking in Table~\ref{tab:scores_benchmark}. For all translations into higher-resource languages, NusaMT-7B scores higher than the GPT models and NLLB-600M, but is either outperformed by NLLB-1.3B or NLLB-3.3B. This could be due to the additional learning that NLLB models transferred from the directions involving similar languages—besides Minangkabau and Balinese—into high-resource languages. However, in translations into Balinese, NusaMT-7B achieves spBLEU scores of 22.1 and 22.9 spBLEU score from English and Indonesian directions, respectively, outperforming all the SoTA models, including the larger NLLB-3.3B by up to +6.69 spBLEU. Similarly, in the Indonesian to Minangkabau direction, NusaMT-7B outperforms NLLB-3.3B by +0.83 spBLEU. Overall, while NusaMT-7B still lags behind SoTAs in translations toward high-resource languages, we observe significant performance gains in translations toward low-resource languages.

\section{Conclusion}
In this paper, we introduced NusaMT-7B, a large language model fine-tuned for low-resource Indonesian languages, with a focus on Balinese and Minangkabau. Our method combines continued pre-training on monolingual data, SFT, and data manipulation techniques using our LLM cleaner and backtranslation. The results from our experiments demonstrate significant performance improvements in translation quality, particularly in directions toward Balinese. Our findings also support the LIMA hypothesis, showing that a smaller, higher-quality dataset can indeed increase model performance. This study presents a promising direction for enhancing machine translation in low-resource settings, contributing to the preservation and revitalization of the many endangered languages in Indonesia and beyond.

\section{Limitations}
There are several limitations to our study. Due to limited GPU resources, we used the Komodo-7B-base model, which constrains our ability to determine the exact number of monolingual tokens each language was pretrained on and prevents us from fully assessing the required content and size of monolingual data for optimal performance. We also did not benchmark against the NLLB-54B Mixture of Experts (MOE) model, NLLB’s largest model \citep{team_2022_no}. In addition, comparisons with models like NLLB are limited by differences in training data, as our model incorporates additional external sources beyond NLLB's SEED and mined bitext datasets. Our findings are also based solely on the spBLEU metric, which may not fully align with translation performance. Furthermore, the ablation study discussed in \ref{bali} was conducted only on language directions involving Balinese; therefore, the performance gains from our chosen techniques may not generalize to other low-resource languages. It is also important to note that, compared to NMT models, our model utilizes significantly more parameters (7 billion), and thus is less computationally efficient during training and inference.

\newpage

{
\small

\bibliographystyle{plainnat}
\bibliography{custom}

}


\newpage

\appendix

\section{Appendix}

\subsection{Prompts}

\subsubsection{Translation Prompt}
\label{translation_prompt}

\begin{verbatim}
Translate this from [source language] to [target language]: 
[source language]: [source]
[target language]:
\end{verbatim}

\textbf{ [Few-shot prompt]}:
\begin{verbatim}
Translate this from English to Balinese: 
English: Astaire continued to act in the 1970s.
Balinese: Astaire sasai maakting ring warsa 1970-an.
\end{verbatim}

\subsubsection{Cleaner Prompt}
\label{datacleaner_prompt}

\begin{verbatim}
You are an expert in aligning and cleaning parallel sentences in different 
languages. You will receive two sentences: one in a source language and 
one in a target language.

Your task is:
1. On the first line, respond with "True" if the sentences have the same 
meaning, otherwise respond with "False".
2. If the first line is "True", provide the cleaned and aligned sentences
on the second and third lines respectively by fixing syntax errors, removing 
noise (such as unnecessary phrases, punctuation or ambiguous 
numbers), and normalizing text (e.g., capitalization).

Here are some examples to guide you:
[Few-shot prompt]

Now, clean the following sentence pairs:
[Batch-prompt]
\end{verbatim}

\textbf{ [Few-shot prompt]}:
\begin{verbatim}
Indonesian: Dengan harga yang bisa dibilang menengah, apa saja yang ditwarkannya?
Balinese: Suratan puniki nénten indik Kabupatén miwah kota ring Kepulauan Riau.

Indonesian: Bahasa daerah memiliki karakteristik yang unik.
Balinese: (32:2) Basa daerah madue "karakteristik" sane soleh.

False

True
Indonesian: Bahasa daerah memiliki karakteristik yang unik.
Balinese: Basa daerah madue karakteristik sane soleh.
\end{verbatim}

\subsection{Filtering}
\label{filtering}
We used OpusFilter, a parallel corpus processing toolkit, \citep{aulamo_2020_opusfilter} to implement several simple filtering methods to remove noisy, low-quality or erroneous parallel sentences.

\textbf{Heuristics.} We apply a few simple heuristics to remove likely noisy sentences. Specifically, we set a length filter between 15 and 500 characters to remove sentences with less than approximately three words and those above the maximum 256 tokens (given an approximate 2.5 characters per token). We also specify a word length ratio of 2 and remove sentences containing words longer than 20 characters, as they often indicate errors. Finally, we deduplicate sentence pairs and remove sentences with excessive punctuation or numerical content beyond a 20\% threshold.

\textbf{Language Identification (LID).}  We intend to preserve only the sentences clearly in the desired source or target language. Thus, we apply the GlotLid V3 LID \citep{hossein_2024_glotlid} on both monolingual and parallel corpora, using only sentences with a language score above a 0.9 threshold, therefore removing sentence pairs that may contain ambiguity and noise.

\textbf{Laser Score.} The LASER3 encoder \citep{team_2022_no} was chosen due to its availability in all the FLORES-200 languages, including Balinese and Minangkabau, as well as its low error rate compared to other multilingual sentence encoders \citep{tan_2023_multilingual}. LASER3 encodes sentences in multiple languages and evaluates the quality of a sentence using xsim \citep{artetxe_2019_marginbased}, as shown in ~\ref{1},

\begin{equation}
\text{score}(x, y) = \text{margin}\left(\cos(x, y), \left(\sum_{z \in \text{NN}_k(x)} \frac{\cos(x, z)}{2k} + \sum_{z \in \text{NN}_k(y)} \frac{\cos(y, z)}{2k}\right)\right)
\label{1}
\end{equation}

 where $x$ denotes the source, $y$ denotes the target sentence embeddings, and $NN_k$ represents the $k$ nearest neighbors of $x$ in other languages. We chose to use the ratio margin function, which is defined as ($ \text{margin}(a,b) = \dfrac{a}{b} $) and set $k=3$. Then, a threshold of $1.09$ was chosen, a slightly higher threshold than NLLB’s $1.06$, since we only want to select high-quality sentence pairs. Bitext with scores lower than this threshold value is filtered out.





\newpage
\section*{NeurIPS Paper Checklist}

\begin{enumerate}

\item {\bf Claims}
    \item[] Question: Do the main claims made in the abstract and introduction accurately reflect the paper's contributions and scope?
    \item[] Answer: \answerYes{}
    \item[] Justification: The abstract and introduction accurately reflect the paper’s contributions, detailing the development of NusaMT-7B for low-resource languages, including specific methods like monolingual pre-training, supervised fine-tuning, and data cleaning, which are all validated by the results presented. The abstract also highlights how the methods can be applied to other languages and datasets.
    \item[] Guidelines:
    \begin{itemize}
        \item The answer NA means that the abstract and introduction do not include the claims made in the paper.
        \item The abstract and/or introduction should clearly state the claims made, including the contributions made in the paper and important assumptions and limitations. A No or NA answer to this question will not be perceived well by the reviewers. 
        \item The claims made should match theoretical and experimental results, and reflect how much the results can be expected to generalize to other settings. 
        \item It is fine to include aspirational goals as motivation as long as it is clear that these goals are not attained by the paper. 
    \end{itemize}

\item {\bf Limitations}
    \item[] Question: Does the paper discuss the limitations of the work performed by the authors?
    \item[] Answer: \answerYes{}
    \item[] Justification: The paper discusses limitations such as the lack of resources for monolingual pre-training, not evaluating more advanced models, and constraints in assessing the optimal size of monolingual data with Komodo-7B-base. Additionally, it reflects on the scope of the claims, referring to the language directions and metrics used, and notes on the model’s computational efficiency in comparison to NMT models.
    \item[] Guidelines:
    \begin{itemize}
        \item The answer NA means that the paper has no limitation while the answer No means that the paper has limitations, but those are not discussed in the paper. 
        \item The authors are encouraged to create a separate "Limitations" section in their paper.
        \item The paper should point out any strong assumptions and how robust the results are to violations of these assumptions (e.g., independence assumptions, noiseless settings, model well-specification, asymptotic approximations only holding locally). The authors should reflect on how these assumptions might be violated in practice and what the implications would be.
        \item The authors should reflect on the scope of the claims made, e.g., if the approach was only tested on a few datasets or with a few runs. In general, empirical results often depend on implicit assumptions, which should be articulated.
        \item The authors should reflect on the factors that influence the performance of the approach. For example, a facial recognition algorithm may perform poorly when image resolution is low or images are taken in low lighting. Or a speech-to-text system might not be used reliably to provide closed captions for online lectures because it fails to handle technical jargon.
        \item The authors should discuss the computational efficiency of the proposed algorithms and how they scale with dataset size.
        \item If applicable, the authors should discuss possible limitations of their approach to address problems of privacy and fairness.
        \item While the authors might fear that complete honesty about limitations might be used by reviewers as grounds for rejection, a worse outcome might be that reviewers discover limitations that aren't acknowledged in the paper. The authors should use their best judgment and recognize that individual actions in favor of transparency play an important role in developing norms that preserve the integrity of the community. Reviewers will be specifically instructed to not penalize honesty concerning limitations.
    \end{itemize}

\item {\bf Theory Assumptions and Proofs}
    \item[] Question: For each theoretical result, does the paper provide the full set of assumptions and a complete (and correct) proof?
    \item[] Answer: \answerNA{} 
    \item[] Justification: The paper is empirical in nature and does not propose new theoretical results or proofs, focusing instead on practical implementation and evaluation of machine translation methods.
    \item[] Guidelines:
    \begin{itemize}
        \item The answer NA means that the paper does not include theoretical results. 
        \item All the theorems, formulas, and proofs in the paper should be numbered and cross-referenced.
        \item All assumptions should be clearly stated or referenced in the statement of any theorems.
        \item The proofs can either appear in the main paper or the supplemental material, but if they appear in the supplemental material, the authors are encouraged to provide a short proof sketch to provide intuition. 
        \item Inversely, any informal proof provided in the core of the paper should be complemented by formal proofs provided in appendix or supplemental material.
        \item Theorems and Lemmas that the proof relies upon should be properly referenced. 
    \end{itemize}

    \item {\bf Experimental Result Reproducibility}
    \item[] Question: Does the paper fully disclose all the information needed to reproduce the main experimental results of the paper to the extent that it affects the main claims and/or conclusions of the paper (regardless of whether the code and data are provided or not)?
    \item[] Answer: \answerYes{} 
    \item[] Justification: The paper provides detailed descriptions of the datasets, training setup, and we release our model and datasets used, making it feasible for others to reproduce the experimental results.
    \item[] Guidelines:
    \begin{itemize}
        \item The answer NA means that the paper does not include experiments.
        \item If the paper includes experiments, a No answer to this question will not be perceived well by the reviewers: Making the paper reproducible is important, regardless of whether the code and data are provided or not.
        \item If the contribution is a dataset and/or model, the authors should describe the steps taken to make their results reproducible or verifiable. 
        \item Depending on the contribution, reproducibility can be accomplished in various ways. For example, if the contribution is a novel architecture, describing the architecture fully might suffice, or if the contribution is a specific model and empirical evaluation, it may be necessary to either make it possible for others to replicate the model with the same dataset, or provide access to the model. In general. releasing code and data is often one good way to accomplish this, but reproducibility can also be provided via detailed instructions for how to replicate the results, access to a hosted model (e.g., in the case of a large language model), releasing of a model checkpoint, or other means that are appropriate to the research performed.
        \item While NeurIPS does not require releasing code, the conference does require all submissions to provide some reasonable avenue for reproducibility, which may depend on the nature of the contribution. For example
        \begin{enumerate}
            \item If the contribution is primarily a new algorithm, the paper should make it clear how to reproduce that algorithm.
            \item If the contribution is primarily a new model architecture, the paper should describe the architecture clearly and fully.
            \item If the contribution is a new model (e.g., a large language model), then there should either be a way to access this model for reproducing the results or a way to reproduce the model (e.g., with an open-source dataset or instructions for how to construct the dataset).
            \item We recognize that reproducibility may be tricky in some cases, in which case authors are welcome to describe the particular way they provide for reproducibility. In the case of closed-source models, it may be that access to the model is limited in some way (e.g., to registered users), but it should be possible for other researchers to have some path to reproducing or verifying the results.
        \end{enumerate}
    \end{itemize}

\item {\bf Open access to data and code}
    \item[] Question: Does the paper provide open access to the data and code, with sufficient instructions to faithfully reproduce the main experimental results, as described in supplemental material?
    \item[] Answer: \answerYes{} 
    \item[] Justification: We release the code for model training on github with specific instructions for the environment and training setup. The data sources are outlined clearly and the compiled parallel sentence dataset is released through huggingface.
    \item[] Guidelines:
    \begin{itemize}
        \item The answer NA means that paper does not include experiments requiring code.
        \item Please see the NeurIPS code and data submission guidelines (\url{https://nips.cc/public/guides/CodeSubmissionPolicy}) for more details.
        \item While we encourage the release of code and data, we understand that this might not be possible, so “No” is an acceptable answer. Papers cannot be rejected simply for not including code, unless this is central to the contribution (e.g., for a new open-source benchmark).
        \item The instructions should contain the exact command and environment needed to run to reproduce the results. See the NeurIPS code and data submission guidelines (\url{https://nips.cc/public/guides/CodeSubmissionPolicy}) for more details.
        \item The authors should provide instructions on data access and preparation, including how to access the raw data, preprocessed data, intermediate data, and generated data, etc.
        \item The authors should provide scripts to reproduce all experimental results for the new proposed method and baselines. If only a subset of experiments are reproducible, they should state which ones are omitted from the script and why.
        \item At submission time, to preserve anonymity, the authors should release anonymized versions (if applicable).
        \item Providing as much information as possible in supplemental material (appended to the paper) is recommended, but including URLs to data and code is permitted.
    \end{itemize}

\item {\bf Experimental Setting/Details}
    \item[] Question: Does the paper specify all the training and test details (e.g., data splits, hyperparameters, how they were chosen, type of optimizer, etc.) necessary to understand the results?
    \item[] Answer: \answerYes{} 
    \item[] Justification: All critical experimental details, such as the data splits, specific hyperparameters (learning rates, batch sizes, lora rank), and loss functions are thoroughly documented.
    \item[] Guidelines:
    \begin{itemize}
        \item The answer NA means that the paper does not include experiments.
        \item The experimental setting should be presented in the core of the paper to a level of detail that is necessary to appreciate the results and make sense of them.
        \item The full details can be provided either with the code, in appendix, or as supplemental material.
    \end{itemize}

\item {\bf Experiment Statistical Significance}
    \item[] Question: Does the paper report error bars suitably and correctly defined or other appropriate information about the statistical significance of the experiments?
    \item[] Answer: \answerNo{} 
    \item[] Justification: The paper does not provide statistical significance measures like error bars or confidence intervals.
    \item[] Guidelines:
    \begin{itemize}
        \item The answer NA means that the paper does not include experiments.
        \item The authors should answer "Yes" if the results are accompanied by error bars, confidence intervals, or statistical significance tests, at least for the experiments that support the main claims of the paper.
        \item The factors of variability that the error bars are capturing should be clearly stated (for example, train/test split, initialization, random drawing of some parameter, or overall run with given experimental conditions).
        \item The method for calculating the error bars should be explained (closed form formula, call to a library function, bootstrap, etc.)
        \item The assumptions made should be given (e.g., Normally distributed errors).
        \item It should be clear whether the error bar is the standard deviation or the standard error of the mean.
        \item It is OK to report 1-sigma error bars, but one should state it. The authors should preferably report a 2-sigma error bar than state that they have a 96\% CI, if the hypothesis of Normality of errors is not verified.
        \item For asymmetric distributions, the authors should be careful not to show in tables or figures symmetric error bars that would yield results that are out of range (e.g. negative error rates).
        \item If error bars are reported in tables or plots, The authors should explain in the text how they were calculated and reference the corresponding figures or tables in the text.
    \end{itemize}

\item {\bf Experiments Compute Resources}
    \item[] Question: For each experiment, does the paper provide sufficient information on the computer resources (type of compute workers, memory, time of execution) needed to reproduce the experiments?
    \item[] Answer: \answerYes{} 
    \item[] Justification: The paper specifies the use of two Nvidia RTX 4090 GPUs and provides estimates of training duration, giving a clear picture of the computational resources required for the experiments.
    \item[] Guidelines:
    \begin{itemize}
        \item The answer NA means that the paper does not include experiments.
        \item The paper should indicate the type of compute workers CPU or GPU, internal cluster, or cloud provider, including relevant memory and storage.
        \item The paper should provide the amount of compute required for each of the individual experimental runs as well as estimate the total compute. 
        \item The paper should disclose whether the full research project required more compute than the experiments reported in the paper (e.g., preliminary or failed experiments that didn't make it into the paper). 
    \end{itemize}
    
\item {\bf Code Of Ethics}
    \item[] Question: Does the research conducted in the paper conform, in every respect, with the NeurIPS Code of Ethics \url{https://neurips.cc/public/EthicsGuidelines}?
    \item[] Answer: \answerYes{} 
    \item[] Justification: The research complies with the NeurIPS Code of Ethics by ensuring responsible use of data, addressing potential biases, and emphasizing the model’s intended use for language preservation without harmful applications.
    \item[] Guidelines:
    \begin{itemize}
        \item The answer NA means that the authors have not reviewed the NeurIPS Code of Ethics.
        \item If the authors answer No, they should explain the special circumstances that require a deviation from the Code of Ethics.
        \item The authors should make sure to preserve anonymity (e.g., if there is a special consideration due to laws or regulations in their jurisdiction).
    \end{itemize}

\item {\bf Broader Impacts}
    \item[] Question: Does the paper discuss both potential positive societal impacts and negative societal impacts of the work performed?
    \item[] Answer: \answerYes{} 
    \item[] Justification: The paper briefly discusses the positive impact on language preservation and cross-cultural communication. However, since this paper focuses on foundational research, a large emphasis is not tied to particular applications.
    \item[] Guidelines:
    \begin{itemize}
        \item The answer NA means that there is no societal impact of the work performed.
        \item If the authors answer NA or No, they should explain why their work has no societal impact or why the paper does not address societal impact.
        \item Examples of negative societal impacts include potential malicious or unintended uses (e.g., disinformation, generating fake profiles, surveillance), fairness considerations (e.g., deployment of technologies that could make decisions that unfairly impact specific groups), privacy considerations, and security considerations.
        \item The conference expects that many papers will be foundational research and not tied to particular applications, let alone deployments. However, if there is a direct path to any negative applications, the authors should point it out. For example, it is legitimate to point out that an improvement in the quality of generative models could be used to generate deepfakes for disinformation. On the other hand, it is not needed to point out that a generic algorithm for optimizing neural networks could enable people to train models that generate Deepfakes faster.
        \item The authors should consider possible harms that could arise when the technology is being used as intended and functioning correctly, harms that could arise when the technology is being used as intended but gives incorrect results, and harms following from (intentional or unintentional) misuse of the technology.
        \item If there are negative societal impacts, the authors could also discuss possible mitigation strategies (e.g., gated release of models, providing defenses in addition to attacks, mechanisms for monitoring misuse, mechanisms to monitor how a system learns from feedback over time, improving the efficiency and accessibility of ML).
    \end{itemize}
    
\item {\bf Safeguards}
    \item[] Question: Does the paper describe safeguards that have been put in place for responsible release of data or models that have a high risk for misuse (e.g., pretrained language models, image generators, or scraped datasets)?
    \item[] Answer: \answerNA{} 
    \item[] Justification: The paper does not release assets that pose high risks for misuse, and therefore, the need for specific safeguards is not applicable.
    \item[] Guidelines:
    \begin{itemize}
        \item The answer NA means that the paper poses no such risks.
        \item Released models that have a high risk for misuse or dual-use should be released with necessary safeguards to allow for controlled use of the model, for example by requiring that users adhere to usage guidelines or restrictions to access the model or implementing safety filters. 
        \item Datasets that have been scraped from the Internet could pose safety risks. The authors should describe how they avoided releasing unsafe images.
        \item We recognize that providing effective safeguards is challenging, and many papers do not require this, but we encourage authors to take this into account and make a best faith effort.
    \end{itemize}

\item {\bf Licenses for existing assets}
    \item[] Question: Are the creators or original owners of assets (e.g., code, data, models), used in the paper, properly credited and are the license and terms of use explicitly mentioned and properly respected?
    \item[] Answer: \answerYes{} 
    \item[] Justification: The paper appropriately credits existing datasets and models, mentioning the sources and licenses where applicable, ensuring proper use and acknowledgment of all assets.
    \item[] Guidelines:
    \begin{itemize}
        \item The answer NA means that the paper does not use existing assets.
        \item The authors should cite the original paper that produced the code package or dataset.
        \item The authors should state which version of the asset is used and, if possible, include a URL.
        \item The name of the license (e.g., CC-BY 4.0) should be included for each asset.
        \item For scraped data from a particular source (e.g., website), the copyright and terms of service of that source should be provided.
        \item If assets are released, the license, copyright information, and terms of use in the package should be provided. For popular datasets, \url{paperswithcode.com/datasets} has curated licenses for some datasets. Their licensing guide can help determine the license of a dataset.
        \item For existing datasets that are re-packaged, both the original license and the license of the derived asset (if it has changed) should be provided.
        \item If this information is not available online, the authors are encouraged to reach out to the asset's creators.
    \end{itemize}

\item {\bf New Assets}
    \item[] Question: Are new assets introduced in the paper well documented and is the documentation provided alongside the assets?
    \item[] Answer: \answerYes{} 
    \item[] Justification: We release our model and compiled dataset through huggingface and included details such as training and dataset sources in the README. The code also provides simple documentation regarding the methods to run the training and validation scripts.
    \item[] Guidelines:
    \begin{itemize}
        \item The answer NA means that the paper does not release new assets.
        \item Researchers should communicate the details of the dataset/code/model as part of their submissions via structured templates. This includes details about training, license, limitations, etc. 
        \item The paper should discuss whether and how consent was obtained from people whose asset is used.
        \item At submission time, remember to anonymize your assets (if applicable). You can either create an anonymized URL or include an anonymized zip file.
    \end{itemize}

\item {\bf Crowdsourcing and Research with Human Subjects}
    \item[] Question: For crowdsourcing experiments and research with human subjects, does the paper include the full text of instructions given to participants and screenshots, if applicable, as well as details about compensation (if any)? 
    \item[] Answer: \answerNA{} 
    \item[] Justification: The paper does not involve crowdsourcing or human subjects, so this consideration does not apply.
    \item[] Guidelines:
    \begin{itemize}
        \item The answer NA means that the paper does not involve crowdsourcing nor research with human subjects.
        \item Including this information in the supplemental material is fine, but if the main contribution of the paper involves human subjects, then as much detail as possible should be included in the main paper. 
        \item According to the NeurIPS Code of Ethics, workers involved in data collection, curation, or other labor should be paid at least the minimum wage in the country of the data collector. 
    \end{itemize}

\item {\bf Institutional Review Board (IRB) Approvals or Equivalent for Research with Human Subjects}
    \item[] Question: Does the paper describe potential risks incurred by study participants, whether such risks were disclosed to the subjects, and whether Institutional Review Board (IRB) approvals (or an equivalent approval/review based on the requirements of your country or institution) were obtained?
    \item[] Answer: \answerNA{} 
    \item[] Justification: The research does not involve human subjects or require IRB approval.
    \item[] Guidelines:
    \begin{itemize}
        \item The answer NA means that the paper does not involve crowdsourcing nor research with human subjects.
        \item Depending on the country in which research is conducted, IRB approval (or equivalent) may be required for any human subjects research. If you obtained IRB approval, you should clearly state this in the paper. 
        \item We recognize that the procedures for this may vary significantly between institutions and locations, and we expect authors to adhere to the NeurIPS Code of Ethics and the guidelines for their institution. 
        \item For initial submissions, do not include any information that would break anonymity (if applicable), such as the institution conducting the review.
    \end{itemize}

\end{enumerate}

\end{document}